\documentclass[sigconf]{acmart}

\AtBeginDocument{%
  }

\copyrightyear{2025}
\acmYear{2025}
\setcopyright{acmlicensed}\acmConference[MM '25]{Proceedings of the 33rd
ACM International Conference on Multimedia}{October 27--31, 2025}{Dublin,
Ireland}
\acmBooktitle{Proceedings of the 33rd ACM International Conference on
Multimedia (MM '25), October 27--31, 2025, Dublin, Ireland}
\acmDOI{10.1145/3746027.3755189}
\acmISBN{979-8-4007-2035-2/2025/10}

\usepackage{algorithm}
\usepackage{algorithmic}
\usepackage{multirow}
\usepackage{enumitem}
\usepackage{xspace}
\usepackage{subfigure}
\newcommand{\model}{PG-Agent\xspace}

\begin{document}

\title{PG-Agent: An Agent Powered by Page Graph}

\author{Weizhi Chen}
\authornote{Both authors contributed equally to this research.}
\affiliation{%
  \department{Zhejiang Key Lab of Accessible Perception \& Intelligent Systems,}
  \institution{Zhejiang University}
  \city{Hangzhou}
  \country{China}}
\email{chenweizhi@zju.edu.cn}
\orcid{0009-0005-1953-1156}

\author{Ziwei Wang}
\authornotemark[1]
\affiliation{%
  \department{Zhejiang Key Lab of Accessible Perception \& Intelligent Systems,}
  \institution{Zhejiang University}
  \city{Hangzhou}
  \country{China}}
\email{wangziwei98@zju.edu.cn}
\orcid{0000-0003-4479-3738}

\author{Leyang Yang}
\affiliation{%
  \department{Zhejiang Key Lab of Accessible Perception \& Intelligent Systems,}
  \institution{Zhejiang University}
  \city{Hangzhou}
  \country{China}}
\email{yangleyang@zju.edu.cn}
\orcid{0009-0006-5553-8037}

\author{Sheng Zhou}
\authornote{Corresponding author.}
\affiliation{%
  \department{Zhejiang Key Lab of Accessible Perception \& Intelligent Systems,}
  \institution{Zhejiang University}
  \city{Hangzhou}
  \country{China}}
\email{zhousheng_zju@zju.edu.cn}
\orcid{0000-0003-3645-1041}

\author{Xiaoxuan Tang}
\affiliation{%
  \institution{Ant Group}
  \city{Beijing}
  \country{China}}
\email{leahxx1226@outlook.com}
\orcid{0009-0008-1765-9924}

\author{Jiajun Bu}
\affiliation{%
  \department{Zhejiang Key Lab of Accessible Perception \& Intelligent Systems,}
  \institution{Zhejiang University}
  \city{Hangzhou}
  \country{China}}
\email{bjj@zju.edu.cn}
\orcid{0000-0002-1097-2044}

\author{Yong Li}
\authornotemark[2]
\affiliation{%
  \institution{Ant Group}
  \city{Hangzhou}
  \country{China}}
\email{liyong.liy@antgroup.com}
\orcid{0009-0005-1664-6425}

\author{Wei Jiang}
\affiliation{%
  \institution{Ant Group}
  \city{Beijing}
  \country{China}}
\email{jonny.jw@antgroup.com}
\orcid{0009-0003-6605-9793}

\renewcommand{\shortauthors}{Weizhi Chen et al.}

\begin{abstract}

Graphical User Interface (GUI) agents possess significant commercial and social value, and GUI agents powered by advanced multimodal large language models (MLLMs) have demonstrated remarkable potential.
Currently, existing GUI agents usually utilize sequential episodes of multi-step operations across pages as the prior GUI knowledge, which fails to capture the complex transition relationship between pages, making it challenging for the agents to deeply perceive the GUI environment and generalize to new scenarios. Therefore, we design an automated pipeline to transform the sequential episodes into page graphs, which explicitly model the graph structure of the pages that are naturally connected by actions. To fully utilize the page graphs, we further introduce Retrieval-Augmented Generation (RAG) technology to effectively retrieve reliable perception guidelines of GUI from them, and a tailored multi-agent framework \model with task decomposition strategy is proposed to be injected with the guidelines so that it can generalize to unseen scenarios.
Extensive experiments on various benchmarks demonstrate the effectiveness of \model, even with limited episodes for page graph construction.
Our codes will be publicly available at \url{https://github.com/chenwz-123/PG-Agent}.

\end{abstract}




\maketitle

\section{Introduction}
\label{sec:intro}

\begin{figure}[t]

  \centering
   \includegraphics[width=0.95\linewidth]{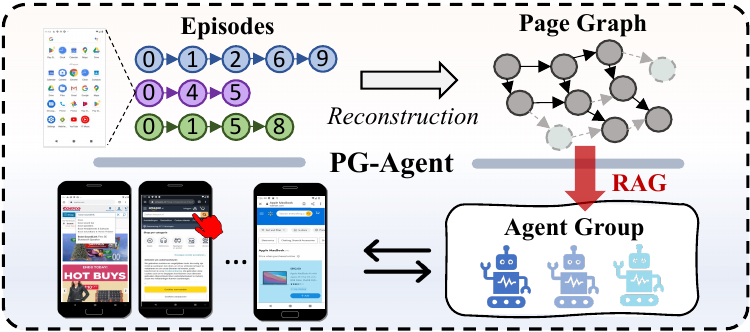}
   \vspace{-0.5em}
   \caption{Illustration of \model. \textbf{(\textit{i})} Convert chain-like episodes into a semantically rich page graph; \textbf{(\textit{ii})} With page graph as GUI prior knowledge, RAG technology assists the tailored multi-agent workflow to enhance GUI navigation.}
   \label{fig:intro_sample}
\vspace{-1.5em}
\end{figure}

The Graphical User Interface (GUI) has become crucial for humans in interacting with mobile devices and websites. 
Recently, there has been a notable increase in interest in GUI agents that can autonomously perform tasks by interacting with the user interface~\cite{nguyen2024guiagentssurvey}. 
It is emerging as a significant topic of study in disciplines such as software engineering and human-computer interaction~\cite{linares2017continuous,ui_tars,llmagentsurvey}, among several others.
Early works employed parsing tools to transform the pages into HTML presentations, utilizing large language models (LLMs) to analyze the page layouts to make decisions~\cite{wu2024copilot,lai2024autowebglm}.
With the rapid development of multimodal large language models (MLLMs)~\cite{qwen2.5vl,gpt4v,internvl,yao2024minicpm,Llama3}, MLLM-based GUI agents become the mainstream architecture, which are able to analyze the screen and generate actions end-to-end.

The GUI agents are conducted in a structured enclosed space where different pages are naturally interconnected through operations like clicking. 
Therefore, it is essential for the agent to possess the awareness of possible actions and their consequent pages.
However, existing works~\cite{mind2web,aitw,odyssey} have collected abundant knowledge from diverse devices, but usually treat them as independent items.
For example, the navigation tasks on GUI involving sequences of multi-step operations across different pages, where each step provides crucial semantics and functionality for the task.
Such linear knowledge restricts the agent to focusing mainly on the consecutive steps and thus lacks the perceptions of possible actions leading to other pages simultaneously. 
As a result, the semantic association information between pages is not fully integrated into the model, leading to significant challenges when the model performs complex tasks, especially on new tasks. 
During the deployment phase, it is probable that a single sequential episode cannot directly instruct the agent to finish the task, while the transition relationships from multiple episodes can provide more clues about possible actions and corresponding results to pave a new way to the target page. 
This raises a natural question: ``How to explicitly model the semantic relationships among pages and enhance the perception capability of GUI Agents in new scenarios?''

Fortunately, the pages of GUI screens naturally form a page graph connected by the actions, and a sequential episode is essentially a path sampling on this graph. 
Inspired by this concept, we can reconstruct the sequential episodes into the page graph, which offers a more comprehensive understanding of the page transitions, rather than the fragmented page connections brought by discrete episodes. Any traversal path in the page graph can be regarded as an effective recombination of original independent knowledge items. Meanwhile, positioning at a node, the agent can easily obtain possible actions from the outgoing edges and perceive consequent pages to assist the navigation process.

Besides, to utilize the page graph as prior knowledge for the agent planning, the Retrieval-Augmented Generation (RAG) technology~\cite{ragfornlptask} is able to effectively leverage it without any parameter modifications, which enables the agent to adapt to different scenarios by simply switching the page graph. In this way, RAG is also able to explicitly retrieve the graph-structured information from the page graph, offering superior semantic perceptions of page transitions. Moreover, the exploration in the page graph of RAG is actually an accessing and integrating process of real actions in episodes, providing an authentic and reliable set of possible actions as guidelines.
Previous works have adopted the RAG to retrieve guidelines like descriptions of widget functions~\cite{li2024appagentv2advancedagent} or reference trajectories in similar tasks~\cite{expel}, which are usually discrete without graph structure to perceive the current scenario deeply. Therefore, a tailored retrieval strategy applied in the page graph is also critical.


%

To tackle the aforementioned challenges, we propose an automated pipeline to transform the sequential episodes into the page graphs, including three stages of page jump determination, node similarity check and page graph update. During the process, we check every action tuple (i.e., the action and the pages before and after it) and gradually update the page graph by combining consecutive in-page operations into one edge and similar pages as one node. Moreover, to retrieving guidelines from the page graphs and fully utilize them, we also design a multi-agent framework \model enhanced by tailored RAG technology. First, we use the summary of the current screen to locate similar nodes in the page graph and conduct breadth first search (BFS) to explore available actions, deriving guidelines like "perform \textit{some actions} can lead to accomplish \textit{some tasks}". With comprehensive guidelines, we divide the reasoning process into several agents following existing works~\cite{wang2024mobile2,mobileagent-E}, incorporate the task decomposition and inject the guidelines into the sub-task planning process, where the perceptions of the GUI scenario are particularly critical. We conduct extensive experiments on three benchmarks and the results demonstrate the effectiveness of PG-Agent, even if we only sample a few episodes to construct the page graph. The primary contributions of this paper can be summarized as follows:



\begin{itemize}[leftmargin=*]

    \item To model the transition relationships between GUI pages, We design an automated pipeline to reconstruct the  discrete episodes into page graphs, which serve as external prior knowledge bases.

    \item We propose \model, a tailored multi-agent framework augmented by RAG technology. With the incorporation of task decomposition, guidelines retrieved from page graphs offer more targeted planning reference.

    \item Experimental results on three benchmarks demonstrate that \model exhibits superior navigation ability with the page graphs. Even if the episodes for page graph construction is limited, the effectiveness remains evident.

\end{itemize}

\begin{figure*}[htb]
  \centering
   \includegraphics[width=1\linewidth]{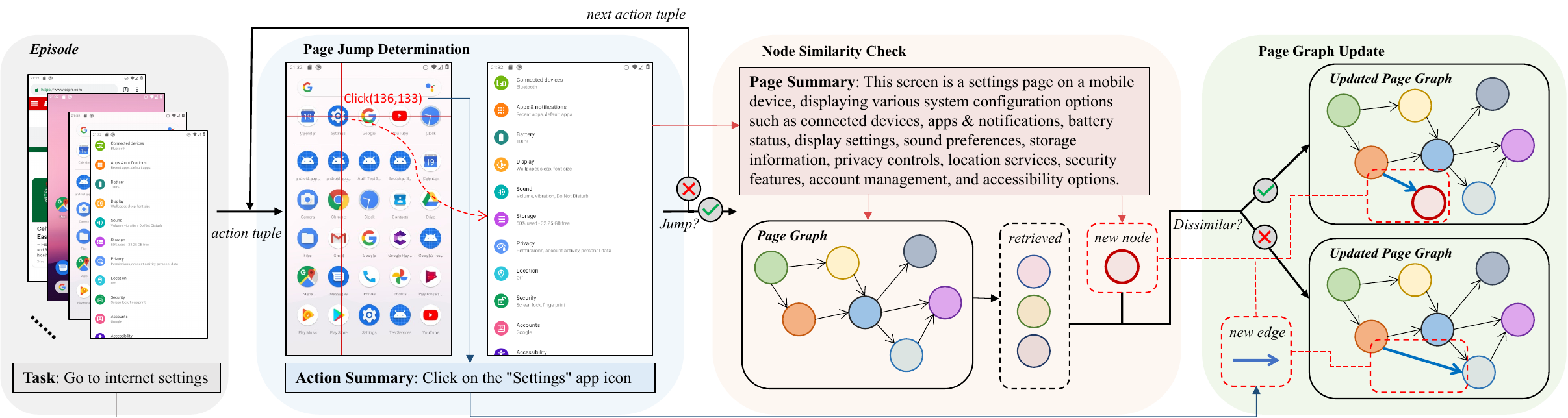}
 \vspace{-1.5em}
   \caption{The overall pipeline of page graph construction. It comprises three stages: page jump determination, node similarity check and page graph update.}
   \label{fig:doc_cons}
    \vspace{-0.8em}
\end{figure*}

\section{Related Work}
\label{sec:relatedwork}

\subsection{Retrieval-Augmented Generation}
Retrieval-Augmented Generation (RAG) can solve the issues of knowledge out-of-distribution in large models, such as output hallucination, lack of domain-specific knowledge, and outdated information by dynamically parsing the input content and retrieving relevant external knowledge~\cite{zhao2024retrievalaugmentedgenerationrag}. Previous research on RAG mainly focused on question-answering related tasks~\cite{rag_min-etal-2024-exploring,rag_shi-etal-2024-generate}, such as using Table-to-Text methods to convert tabular data into textual form to enhance the document QA capability of LLM~\cite{rag_min2024exploringimpacttabletotextmethods}, using multi-modal embedding technology to uniform knowledge of different modalities to enhance multi-modal QA of the foundational model~\cite{hu2023revealretrievalaugmentedvisuallanguagepretraining,long2024generativemultimodalknowledgeretrieval}, and utilizing the chunking methods to truncate the query and realize multi-granularity retrieval of external knowledge~\cite{chen-etal-2024-dense,yepes2024financialreportchunkingeffective}. Given that graph structures effectively represent complex data relationships and enable efficient information retrieval, the application of RAG technology to graph data has emerged as an interesting research focus~\cite{peng2024graphretrievalaugmentedgenerationsurvey}. GraphRAG~\cite{graph_rag_edge2025localglobalgraphrag} adopts a clustering approach by connecting small text blocks via semantic similarity, then applying community detection to group nodes, and finally summarizing query answers by analyzing node community responses. To model document relationships: Munikoti et al.~\cite{munikoti2023atlanticstructureawareretrievalaugmentedlanguage} developed a heterogeneous document graph capturing multiple document-level relations, while Li et al.~\cite{li2024graphneuralnetworkenhanced} and Wang et al.~\cite{wang2023knowledgegraphpromptingmultidocument} established passage-level connections based on shared keywords. In the mobile agent scenario, there are also some works that use RAG to enhance the base model by providing additional interaction knowledge~\cite{li2024appagentv2advancedagent,expel}. However, they treat the traffic data as independent trajectory chains. We argue that in GUI scenarios, the data formed by the jump relationship between different screen pages is a global graph structure rather than discrete chains. Thus, ignoring the structured signals between pages will limit the model's knowledge learning in this domain.

\subsection{GUI Agent}

Recent progress has begun to adopt LLMs~\cite{yao2023react, reflexion, expel} to build autonomous agents, leveraging LLMs' extensive world knowledge and strong reasoning capabilities for task planning and execution to achieve human-like capabilities. Structural text replaces the original GUI image input into the LLMs. 
With the emergence of MLLMs, visual signals of images are projected into natural language space. Therefore, existing research tends to directly use MLLMs to build agents, so as to naturally process the visual information in the GUI field.
One notable approach is to use large-scale general models, such as GPT-4v~\cite{gpt4v}, as GUI agents. Many studies use prompt engineering to guide these models to perform complex tasks. AppAgent~\cite{appagent} is built on GPT-4v, generating guidance documents through exploration phase to assist decision-making. Mobile-Agent-v2~\cite{wang2024mobile2} first proposes multi-agent collaboration in GUI scenarios to improve the decision-making effect of each step. 
Another research direction focuses on fine-tuning smaller MLLMs~\cite{seeclick, showui, cogagent} using GUI-specific datasets to bridge the domain gap between common images and GUI screens. 

\begin{figure*}[t]
  \centering
   \includegraphics[width=0.8\linewidth]{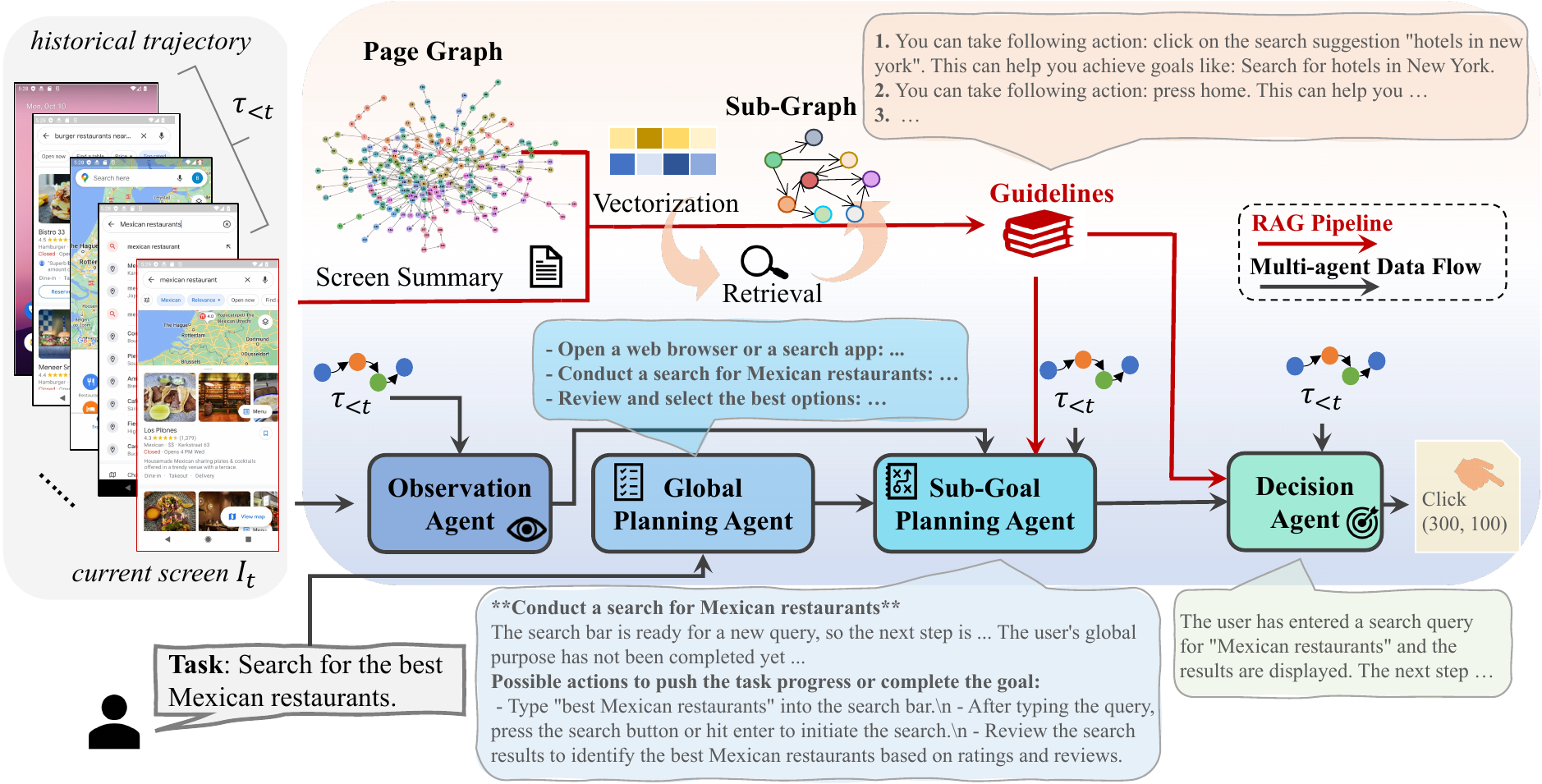}
   \caption{The framework of multi-agent workflow. It comprises two parts: RAG pipeline and multi-agent group.}
   \label{fig:multi_agent_workflow}
   \vspace{-0.5em}
\end{figure*}

\section{Method}
\label{sec:method}

In this section, we will illustrate how to transform chained action episodes into structured page graphs, along with readable guideline documents. Subsequently, we introduce \textbf{\model} that is tailored to leverage the page graphs to achieve precise GUI navigation.
\subsection{Page Graph Construction}\label{page_graph_construction}

Naturally, the pages and their links within a website or an application form a graph structure, and an episode to complete a navigation task actually represents a walking path in this graph. Thus, with existing episodes in a specific GUI scenario where relevant websites or applications are limited, we can construct the corresponding page graph as future guidance in this scenario. We design our pipeline purely based on visual clues without additional modal inputs such as the page's DOM or HTML architecture. The overall pipeline for page graph construction is shown in Figure~\ref{fig:doc_cons}, including three stages: page jump determination, node similarity check and page graph update.

\textbf{Page Jump Determination.} Assuming an action tuple in the episode $E$ with task $T$ is $(I_{before},A,I_{after})$, where $I_{before}$ and $I_{after}$ represent the screenshot images before and after the action $A$, respectively. First, the actions need to be converted into natural language. For actions involving specific coordinates, their meanings will be lost when separated from the corresponding images. Therefore, a MLLM~\cite{qwen2.5vl} is used to summarize them:
\begin{equation} 
    S_{action}=\textrm{MLLM}(I_{before},A,\mathbb{P}_{action}),
\end{equation}
where $S_{action}$ is the summary of the action and $\mathbb{P}_{action}$ is the prompt for action summary. Then, considering fewer redundant nodes and retrieval effectiveness, only unique pages are adopted to build the page graph. Therefore, it is first necessary to determine whether the action $A$ triggers the page jump.
\begin{equation}
Y_{jump}=\textrm{MLLM}(I_{before},I_{after},S_{action},\mathbb{P}_{jump}),
\end{equation}
where $Y_{jump}\in \textrm{[Yes, No]}$ represents the determination result and $\mathbb{P}_{jump}$ is the prompt template for page jump assessment. If the result is "No", the action is usually in-page operation like typing words or opening a drawer that do not lead to new pages, we will store this action summary $S_{action}$ in a queue $Q_{action}$ and directly process the next action tuple in the episode.

\textbf{Node Similarity Check.} When the result $Y_{jump}$ is "Yes", it means this action successfully lead to different page. Then we take the following step to summarize the image after action based on the overall function of the page and key components displayed for node similarity check:
\begin{equation} 
    S_{page}=\textrm{MLLM}(I_{after},\mathbb{P}_{page}),
\end{equation}
where $S_{page}$ denotes the summary of the page and $\mathbb{P}_{page}$ is the template for page summary processing. Next, a dual-level similarity check from semantic aspect and pixel aspect was carried out. From semantic aspect, we employ similarity search with retrieval model to retrieve the top-$n$ most similar page summaries from nodes of page graph $\mathcal{G}=(\mathcal{N},\mathcal{V})$, which is empty initially:
\begin{equation} 
    {S_{node}}=\textrm{Retrieval}(S_{page},\mathcal{G}),
\end{equation}
where ${S_{node}}=[S_1,S_2,\dots,S_n]$ is composed of retrieved page summaries. Subsequently, we use the MLLM to further select the index of the most similar one:
\begin{equation} 
    id=\textrm{MLLM}(I_{after},{S_{node}},\mathbb{P}_{select}),
\end{equation}
where $id\in [1,2,\dots,n]$ and $\mathbb{P}_{select}$ is the prompt for index selection. From pixel aspect, we extract original image $I_{id}$ corresponding to selected page summary and compare it with image $I_{after}$ to finally conclude whether the page can pass the node similarity check: 
\begin{equation}\label{equa:dissimilar}
Y_{dissimilar}=\textrm{MLLM}(I_{after},I_{id},\mathbb{P}_{dissimilar}),
\end{equation}
where $Y_{dissimilar}\in \textrm{[Yes, No]}$ is the check result and $\mathbb{P}_{dissimilar}$ is the prompt for similarity check.

\textbf{Page Graph Update.} When the result $Y_{dissimilar}$ is "Yes", it means that this new page is unique enough among existing nodes in the page graph $\mathcal{G}$, and we will create a new node $\mathcal{N}_{new}=(S_{after},L_{after})$ to represent the image $I_{after}$, where $L_{after}$ is the image location of $I_{after}$. Image location $L$ will only be utilized in Equation~\ref{equa:dissimilar} to get original image $I_{id}$, so the final page graph will not contain the pixel information of images, avoiding the huge space occupancy of page graph. Besides, the node representing the image $I_{before}$ can be formulated as $\mathcal{N}_{before}$. Following, we incorporate the action summary $S_{action}$ into stored action queue $Q_{action}$ and combime it with the task description $T$ of the episode as a new edge $\mathcal{E}_{new}=(Q_{action},T)$. In this way, we can guide the agent to follow these actions when completing similar tasks. Next, we insert directed tuple $(\mathcal{N}_{before},\mathcal{E}_{new},\mathcal{N}_{new})$ into page graph $\mathcal{G}$ to finish update:
\begin{equation}
\mathcal{G}=\mathcal{G} \cup (\mathcal{N}_{before},\mathcal{E}_{new},\mathcal{N}_{new}).
\end{equation}
If the decision of similarity check is ``No", page $I_{after}$ can actually be represented by existing node $N_{id}$ of image $I_{id}$, so the tuple to be inserted will be changed to $(\mathcal{N}_{before},\mathcal{E}_{new},\mathcal{N}_{id})$:
\begin{equation}
\mathcal{G}=\mathcal{G} \cup (\mathcal{N}_{before},\mathcal{E}_{new},\mathcal{N}_{id}).
\end{equation}

\subsection{Multi-agent Workflow}\label{sec:mluti_agent_execution}

The workflow of agent framework could be formalized as a Markov Decision Process (MDP)~\cite{yao2023react,reflexion,chen2024automanual}. Previous work mainly use a LLM, such as GPT-4~\cite{gpt4}, to structure image text with the help of additional parsing tools~\cite{fu2024autoguide,expel}, or a separate MLLM~\cite{appagent,lu2024omniparser}, such as GPT-4V~\cite{gpt4v}, to preprocess the image using the Set-of-Marks strategy~\cite{yang2023setofmark}. Then the design of the agent workflow is completed based on the prompt engineering. However, under the warper paradigm, the long-content poses a challenge to the reasoning performance of the model, making the model at risk of being "lost-in-the-middle"~\cite{liu2023lostmiddlel}. 

In this section, we adopt the multi-agent workflow that logically connects multiple MLLM-based agents with different roles. Each agent receives different input content and only completes specific tasks, which alleviates the context processing pressure of the model, and then spends more efforts on the task reasoning stage. Based on this architecture proposed before~\cite{wang2024mobile2}, we further introduce the task decomposition concept into agent group.  

Our multi-agent workflow is shown in Figure~\ref{fig:multi_agent_workflow} and it mainly consists of two key parts: (1) RAG pipeline, which retrieves helpful guidelines from page graph based on screen status; (2) Multi-agent group: agents with different roles, i.e., global planning agent, observation agent, sub-task planning agent and decision agent.

\textbf{Guidelines Retrieval.} The guidelines retrieved from the page graph are the core mechanism to enhance the generalization capability of agents in new GUI scenarios. First, we prompt the MLLM to analyze the current screen status $I_t$ and generate a screen state description $S_{I_t}$. Subsequently, $S_{I_t}$ is vectorized to retrieve the top $n$ most similar nodes $\mathcal{N}$ from page graph $\mathcal{G}$:
\begin{equation} 
   S_{I_t} =\textrm{MLLM}(I_{t}, \mathbb{P}_{sum}),
\end{equation}
\begin{equation} 
   \mathcal{N} =\textrm{Retrieval}(S_{I_t}, \mathcal{G}),
\end{equation}
where $\mathbb{P}_{sum}$ represents the template for screen summary. Then we extract the action queues $Q$ stored in the outgoing edges $\mathcal{E}$ of node set $\mathcal{N}$. Besides, starting from every outgoing edge $\mathcal{E}_i$, we conduct BFS with $l$ layers and gather the tasks stored in the explored edges:
\begin{equation} 
   T_{i} =\textrm{BFS}(\mathcal{E}_{i}, \mathcal{G}),
\end{equation}
where $T_i$ is the gathered achievable tasks from the edge $\mathcal{E}_{i}$. We combine the action queue and achievable tasks as the guidelines:
\begin{equation} \label{gl_retrivel}
   {G}_{I_t} =[(Q_1,T_1),(Q_2,T_2),\dots,(Q_k,T_k)],
\end{equation}
where $k$ is the number of retrieved guidelines. Each tuple $(Q_i,T_i)$ donates that the agent could follow the action queue $Q_i$ to complete tasks recorded in set $T_i$.

\textbf{Global Planning Agent.} $\mathcal{P}^{G}_{agent}$ is used to perform a global high-level sub-task decomposition of the user's task ${T}_g$, breaking down the complex task into relatively simple and abstract sub-tasks (i.e., the global plan). In this way, the guidelines can inspire the agent to focus on completing the current sub-task. This process can formulated as:
\begin{equation}\label{eq_global}
\mathcal{R}_g = \mathcal{P}^{G}_{agent}(I_t, T_g).
\end{equation}

\textbf{Observation Agent.} $\mathcal{O}_{agent}$ is responsible for transforming the pixel information into textual perceptions. It observes the screen and provides useful visual clues along with a high-level abstract functional description. In this stage, we introduce the historical interaction record $\tau_{<t}$ from the previous moment to help $\mathcal{O}_{agent}$ perceive task progress. With user's task ${T}_g$, $\mathcal{O}_{agent}$ can be formulated as:

\begin{equation}\label{eq_observation}
\mathcal{R}_{o} = \mathcal{O}_{agent}(I_t,{T}_g,\tau_{<t}),
\end{equation}
where $\tau_{<t} = (I_0,a_0, I_1,a_1,...,I_{t-1},a_{t-1})$ and $a_t$ represents the action executed at time-step $t$.
The goal of the $\mathcal{O}_{agent}$ is to directly provide explicit screen details to the decision agent $\mathcal{D}_{agent}$, so that it can pay more attention in reasoning.


    


\textbf{Sub-Task Planning Agent.} $\mathcal{P}^{S}_{agent}$ selects a sub-task that matches the current screen state from the global plan $\mathcal{R}_g$, provides a detailed description of the current task suggestion, and generates a candidate action list. Based on screen observation $\mathcal{R}_o$, global plan $\mathcal{R}_g$, retrieved guidelines $G_{I_t}$, and historical trajectory $\tau_{<t}$, this process can be formulated as:
\begin{equation} \label{eq_subgoal}
\mathcal{R}_s = \mathcal{P}^{S}_{agent}(I_t, \mathcal{R}_o, \mathcal{R}_g, G_{I_t}, \tau_{<t}).
\end{equation}

\textbf{Decision Agent.} $\mathcal{D}_{agent}$ eventually chooses the specific action to be performed in the current screen state $I_t$ from the candidate action list $\mathcal{R}_s$ via analyzing the previously generated content. The decision process can be formulated as:
\begin{equation} \label{eq_decison}
\mathcal{R}_{d} = \mathcal{D}_{agent}(I_t, \mathcal{R}_{o}, \mathcal{R}_s,G_{I_t}, \tau_{<t}).
\end{equation}

As shown in Figure~\ref{fig:multi_agent_workflow}, when given a screen image $I_t$, \textbf{RAG pipeline} will summarizes $I_t$, vectorizes\cite{douze2024faiss} the summary to retrieve similar nodes from page graph $\mathcal{G}$ and explore them to generates guidelines $G_{I_t}$. Then, \textbf{Observation Agent} $\mathcal{O}_{agent}$ will carefully perceive the screen status $I_t$ and produce detailed description of the page. Next, \textbf{Global Planning Agent} $\mathcal{P}^{G}_{agent}$ will decouple user's global task $T_g$ into several clear, coherent and relatively easy sub-tasks $\mathcal{R}g$. Afterwards, \textbf{Sub-Task Planning Agent} $\mathcal{P}^{S}_{agent}$ will conduct in-depth analysis of the context information, including $I_t$, $\mathcal{R}_{o}$, $\mathcal{R}_g$, $G_{I_t}$ and $\tau_{<t}$, and complete the fine-grained plan $\mathcal{R}s$ of the current sub-task under the help of guidelines $G_{I_t}$. Finally, the \textbf{Decision Agent} $\mathcal{D}_{agent}$ will use $\mathcal{R}_{o}$, $\mathcal{R}_s$, $G_{I_t}$, and $\tau_{<t}$ to generate the final decision $\mathcal{R}_d$ to predict the action that should be performed in the current state to advance the task $T_g$.

For more details on page graph and multi-agent workflow, please refer to the \textbf{Supplementary Material}.

\section{Experiment}

\begin{figure}[t]
  \centering
  
   \includegraphics[width=1.0\linewidth] 
   {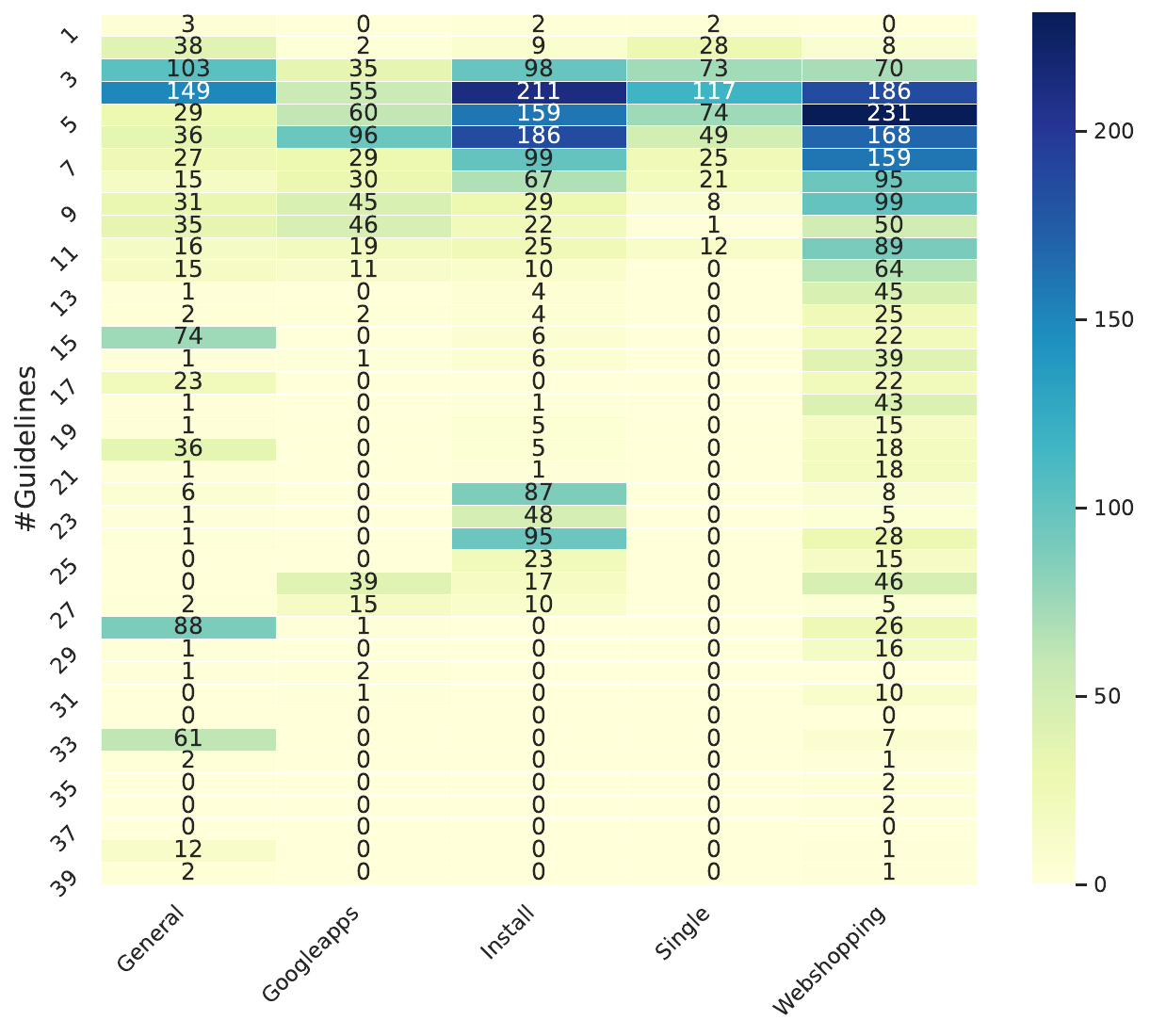}
   \vspace{-1.1em}
   \caption{The data distribution of guidelines in AITW dataset. The x-axis represents the scenario category and the y-axis represents the number of retrieved guidelines at each step.}
   \label{fig:ref_aitw}
   \vspace{-0.5em}
\end{figure}

\subsection{Experimental Setting}\label{sec:exp_setting}

\textbf{Benchmark Dataset.} To assess the navigation ability in both mobile and website environment, we evaluate our \model on two GUI agent datasets: Android in the Wild (AITW)~\cite{aitw}, Mind2Web~\cite{mind2web} and GUI Odyssey~\cite{odyssey}:

\begin{table*}[htbp]
\caption{Comparison of \model with different methods on Mind2Web dataset. The best and second-best results in each column are highlighted in \textbf{bold} font and \underline{underlined}.}

    \centering
    \resizebox{0.8\linewidth}{!}{
    \begin{tabular}{l|ccccccccc}
        \toprule
         \multirow{2}{*}{Method} &  \multicolumn{3}{c}{{Cross-Task}} & \multicolumn{3}{c}{Cross-Website} & \multicolumn{3}{c}{{Cross-Domain}} \\ 
        \cmidrule{2-10} 
        &  {Ele.Acc}&{Op.F1}&{Step SR}&{Ele.Acc}&{Op.F1}&{Step SR}& {Ele.Acc}&{Op.F1}&{Step SR}\\ 
        \midrule
    
MindAct         & \underline{55.1}          & 75.7          & \underline{52.0}          & \underline{42.0}          & 65.2          & 38.9          & 42.1          & 66.5 & 39.6          \\
GPT-4V          & 46.4          & 73.4          & 40.2          & 38.0          & 67.8          & 32.4          & 42.4          & 69.3 & 36.8          \\
Qwen2.5-VL-72B          & 31.9          & 84.6          & 26.2          & 35.7          & 80.7          & 27.9          & 32.0          & 83.2 & 25.0          \\
OmniParser     & 42.4          & \textbf{87.6} & 39.4          & 41.0          & \textbf{84.8} & 36.5          & 45.5          & \textbf{85.7} & \underline{42.0}          \\
    \midrule
   \model & \textbf{59.0} & \underline{84.7}          & \textbf{52.9} & \textbf{57.3} & \underline{81.2}          & \textbf{48.7} & \textbf{60.2} & \underline{84.5} & \textbf{53.3}\\
        \bottomrule
    \end{tabular}
    }
 \vspace{0.2em}
    
    \label{tab:main_res_mind2web}
\end{table*}
\begin{table*}[tbp]\LARGE
\caption{Comparison of \model with different methods on GUI Odyssey dataset. The best and second-best results in each column are highlighted in \textbf{bold} font and \underline{underlined}.}

\centering
\resizebox{0.68\linewidth}{!}{
    \begin{tabular}{l|cccccc|c}
    \toprule
    Method & Tool  & Information& Shopping& Media& Social& Multi-Apps& Overall\\
    \midrule
GeminiProVision&  3.3            & 4.0          & 2.3           & 4.3          & 1.5           & 3.2           & 4.9              \\
CogAgent       & 11.8           & 15.7         & 10.7          & 9.2          & 11.7          & 13.1          & 10.7             \\
GPT-4V         & 18.8           & 23.5         & 20.2          & 19.2         & 16.9          & 13.8          & 19.0             \\
GPT-4o         & 20.4           & 20.8         & 16.3          & 31.9         & 15.4          & 21.3          & 16.7             \\
Qwen2.5-VL-72B     & \underline{46.6}  & \underline{60.0}  & \underline{44.0}   & \underline{32.4}& \underline{46.1}   & \textbf{54.6 }         & \underline{42.4}      \\ 
\midrule
\model & \textbf{48.6}  & \textbf{61.5} & \textbf{47.2} & \textbf{35.5} & \textbf{46.9} & \underline{52.6} & \textbf{47.7}  
    \\
    \bottomrule
    \end{tabular}
    }

\vspace{0.2em}
    \label{tab:main_res_odyssey}
\end{table*}

\begin{itemize}[leftmargin=*]

\item \textbf{AITW}: The episodes in AITW dataset are collected in Android mobile phone environment, which are divided into five scenarios: General, Install, GoogleApps, Single, and WebShopping. We follow the split setting of SeeClick~\cite{seeclick}. For simplicity, we randomly sample 1/10 episodes from training split to construct concise page graphs based on different scenarios. The specific statistics are shown in Table~\ref{tab:aitw_graph_data}.

\item \textbf{Mind2Web}: Mind2Web dataset contains over 2,000 open-ended tasks collected from 137 real websites, covering five scenarios: Entertainment, Travel, Shopping, Service, and Info. Also, we randomly sample episodes from the training set. The benchmark on Mind2Web is not divided by scenarios, but categorized into cross-task, cross-website, and cross-domain to test the generalization ability of the agent. Therefore, the Service and Info scenarios that only appear at cross-domain test do not have corresponding page graphs, so we will use the page graphs of other scenarios during evaluation.

\item  \textbf{GUI Odyssey}: GUI Odyssey dataset is designed to evaluate the navigation ability of the agent in cross-app tasks. This dataset contains more than 7,000 episodes with an average of 15+ steps, including 6 different scenarios from 201 apps. Similarly, we sample part of training episodes of GUI Odyssey to build page graphs.

\end{itemize}
\noindent The specific statistics of episodes sampling of dataset Mind2Web and GUI Odyssey are listed in supplementary materials.

\noindent  \textbf{Model.} In this paper, we adopt BGE-M3~\cite{bge-m3} as our vectorization model and FAISS technique~\cite{douze2024faiss} for similarity retrieval. Besides, we utilize Qwen2.5-VL-72B~\cite{qwen2.5vl} as our base MLLM considering its strong ability at understanding GUI screens.

\noindent  \textbf{Hyperparameters.} According to the statistics of retrieved guidelines (\textit{GL}), as shown in Figure.\ref{fig:ref_aitw}, we set the maximum number of \textit{GL} (Equation~\ref{gl_retrivel}) to 20 for AITW dataset. Besides, we set 20 for GUI Odyssey and 10 for Mind2Web, whose distribution of \textit{GL} in different scenarios can be seen in the supplementary materials. Besides, we set the maximum number of layers $l$ for BFS to 3 and the number of retrieved nodes of page similarity search $n$ to 4.

\subsection{Main Result}\label{sec:main_res}

\textbf{AITW.} We follow the setting of AITW to calculate the action matching score as the metric. As shown in Table~\ref{tab:main_res_aitw}, \model yields the best average performance compared to current API-based agents. Among the scenarios, the action accuracy in GoogleApps is the most prominent, which exceeds state-of-art result by 13.4\%. The result demonstrates that graph RAG technique can help API-based agent improve the execution accuracy in scenarios where prior knowledge is available. For error cases in Single scenario, we find the length of these episodes is short and the step where the task is supposed to end has ambiguity, i.e., our agent tends to continue executing some actions to completely finish the task. We further analyze this situation in the {Supplementary Materials}.

\begin{table}[htbp]\LARGE
\caption{Comparison of \model with different methods on AITW dataset. The best and second-best results in each column are highlighted in \textbf{bold} font and \underline{underlined}.}
\centering
\resizebox{1.0\linewidth}{!}{
    \begin{tabular}{l|ccccc|c}
    \toprule
    Method             & General       & Install       & G.Apps        & Single        & WebShop.       & Overall       \\
    \midrule
ChatGPT-CoT        & 5.9           & 4.4           & 10.5          & 9.4           & 8.4           & 7.7           \\
PaLM2-CoT          & -             & -             & -             & -             & -             & 39.6          \\
GPT-4V             & 41.7          & 42.6          & 49.8          & \underline{72.8}    & 45.7          & 50.5          \\
Qwen2.5-VL-72B        & 35.9           & \underline{58.5}           & \underline{58.8}          & 50.7           & 36.6           & 48.1           \\
OmniParser         & \underline{48.3}    & 57.8    & 51.6    & \textbf{77.4} & \underline{52.9}    & \underline{57.5}    \\
\midrule
\model & \textbf{51.9} & \textbf{62.4} & \textbf{65.0} & 64.7          & \textbf{53.7} & \textbf{59.5}   
    \\
    \bottomrule
    \end{tabular}
    }\
    \vspace{0.2em}
    
    \label{tab:main_res_aitw}
\end{table}

\noindent \textbf{Mind2Web.} In Mind2Web dataset, we calculate element accuracy (Ele.Acc), operation f1 (Op.F1) and step success rate (Step SR) as the metrics. Results in Table~\ref{tab:main_res_mind2web} show that \model achieves the optimal performance in both Ele.Acc and Step.SR metric, and the second-best in Op.F1. Besides, the significant improvements can be observed in cross-domain split, where we lack for relevant prior knowledge in Service and Info scenarios. This indicates that the episodes from other scenarios also provide valuable reference, which proves the generality of constructed page graph.

\begin{table*}[]
\caption{Ablation results on Mind2Web. The best and second-best results in each column are highlighted in \textbf{bold} font and \underline{underlined}. \textit{GL}, STP-Agent, D-Agent are the abbreviations of guidelines, sub-task planning agent and decision agent respectively.}

    \centering
    \resizebox{0.8\linewidth}{!}{
    \begin{tabular}{l|ccccccccc}
        \toprule
          \multirow{2}{*}{Method}&  \multicolumn{3}{c}{{Cross-Task}} & \multicolumn{3}{c}{Cross-Website} & \multicolumn{3}{c}{{Cross-Domain}} \\ 
        \cmidrule{2-10} 
        &  {Ele.Acc}&{Op.F1}&{Step SR}&{Ele.Acc}&{Op.F1}&{Step SR}& {Ele.Acc}&{Op.F1}&{Step SR}\\ 
        \midrule
    \model & \underline{59.0}	& \textbf{84.7}	& \textbf{52.9}	& 57.3	& 81.2	& \textbf{48.7}	& \textbf{60.2}	&\textbf{ 84.5}	&\textbf{ 53.3} \\
    \midrule
    \textit{w/} \textit{GL} in D-Agent & 58.1	& 84.0	& 50.7	& \underline{57.5}	& \underline{82.0}	& \underline{48.5} &\underline{59.9}	&82.7	&51.5 \\
    \textit{w/} \textit{GL} in STP-Agent & \textbf{59.4}	& \underline{84.5} &	\underline{52.4} &	56.9 &	\textbf{83.1}	& 48.0 & 59.5&	\underline{83.4}	&\underline{52.1}  \\
   \textit{w/o} \textit{GL} & 58.9&	82.8 &	50.2 &	\textbf{57.6} &	80.6 &	47.6 &	59.4 &	81.3 &	50.4\\
        \bottomrule
    \end{tabular}
    }
 \vspace{0.2em}
    
    \label{tab:ablation_res_mind2web}

\end{table*}

\begin{table*}[]\small
\caption{Ablation results of guidelines for different actions on Mind2Web. The metric is Op.F1 value, where the best result is highlighted in \textbf{bold}, and \textit{w/o GL} means removing the RAG pipeline.}

    \centering
    \resizebox{0.7\linewidth}{!}{
    \begin{tabular}{l|ccccccccc}
        \toprule
          \multirow{2}{*}{Method}&  \multicolumn{3}{c}{{Cross-Task}} & \multicolumn{3}{c}{Cross-Website} & \multicolumn{3}{c}{{Cross-Domain}} \\ 
        \cmidrule{2-10} 
        &  {\tt CLICK}&{\tt SELECT}&{\tt TYPE}&{\tt CLICK}&{\tt SELECT}&{\tt TYPE}&{\tt CLICK}&{\tt SELECT}&{\tt TYPE}\\ 
        \midrule
    \model & {90.9}	&\textbf{31.2}	&\textbf{49.5}&	{88.0}&	\textbf{47.3}&	\textbf{44.0}&	{88.8}	&\textbf{47.1}	&\textbf{56.3}\\
   \textit{w/o} \textit{GL} &\textbf{92.9}&	29.8&	29.6&	\textbf{91.0}&	40.5&	26.0&	\textbf{89.6}&	40.6&	31.7 \\
        \bottomrule
    \end{tabular}
    }

     \vspace{0.2em}
    \label{tab:ablation_action}
\end{table*}

\noindent \textbf{GUI Odyssey.} For GUI Odyssey, we adhere the original metric setting~\cite{odyssey} . From the results in Table~\ref{tab:main_res_odyssey}, we can observe \model produces the best results in most scenarios and surpasses other API-based agents, while only in Multi-Apps scenario it is suboptimal. This demonstrates that the guidelines retrieved from page graph cast some insights into unfamiliar scenario for the agent and actually improve the planning process and execution process during the navigation.

\subsection{Ablation Study}
In our \model, as shown in Figure~\ref{fig:multi_agent_workflow}, our graph RAG pipeline extracts relevant guidelines (\textit{GL}) from the page graph (Section~\ref{page_graph_construction}) and acts in the planning stage (Sub-Task Planning Agent) and decision stage (Decision Agent) respectively. In this section, we use the control variable method to analyze the impact of \textit{GL} on \model. Specifically, we prompt the retrieved \textit{GL} to different agents, and the results are shown in Table~\ref{tab:ablation_res_aitw} and Table~\ref{tab:ablation_res_mind2web}. It can be seen that on AITW\cite{aitw}, our \model achieves the best results, while removing \textit{GL} (\textit{w/o GL}) leads to a general decline in performance.

\begin{table}[htbp]\LARGE
\caption{Ablation results on AITW. The best and second-best results in each column are highlighted in \textbf{bold} font and \underline{underlined}. \textit{GL}, STP-Agent, D-Agent are the abbreviations of Guidelines, Sub-Task Planning Agent and Decision Agent respectively.}
\centering
\resizebox{\linewidth}{!}{
    \begin{tabular}{l|ccccc|c}
    \toprule
    Method             & General       & Install       & G.Apps        & Single        & WebShop.       & Overall       \\
    \midrule
    \model & \textbf{51.9} & \textbf{62.4} & \textbf{65.0} & 64.7 & \textbf{53.7} & \textbf{59.5} \\
    \midrule
    \textit{w/} \textit{GL} in D-Agent & 50.8	& \underline{60.5}	& \underline{63.8}	& \textbf{66.6}	& \underline{53.4}	& \underline{59.0}\\
    \textit{w/} \textit{GL} in STP-Agent & \underline{51.4}	& 59.5 &	62.8 &	\underline{66.1}	& 52.8	& 58.5 \\
   \textit{w/o} \textit{GL} & 50.0	& 59.8	& 63.4	& 65.4	& 52.7 &	58.3 \\
    \bottomrule
    \end{tabular}
    }
    \vspace{0.2em}
    
    \label{tab:ablation_res_aitw}
\end{table}

\begin{table*}[t]
\caption{The impact of the page graph on \model constructed with different data of episodes, where random sampling follows the setting in Section~\ref{sec:exp_setting}, while full episodes means we utilize all episodes from the training set.}
    \centering
    \resizebox{0.8\linewidth}{!}{
    \begin{tabular}{l|ccccc|ccc}
        \toprule
          \multirow{2}{*}{Data Source}&  \multicolumn{5}{c|}{AITW} & \multicolumn{3}{c}{Mind2Web} \\ 
        \cmidrule{2-9} 
              & General       & Install       & G.Apps        & Single        & WebShop. & Cross-Task & Cross-Website & Cross-Domain\\ 
        \midrule
    Random Sampling & \textbf{51.9} & \textbf{62.4} & \textbf{65.0} & \textbf{64.7}          & 53.7 & \textbf{52.9} & 48.7 & 53.3\\
   Full Episodes &50.5 & 59.5 & 63.2 & 61.4          & \textbf{54.6} & 52.6 & \textbf{50.0} & \textbf{54.0}\\
        \bottomrule
    \end{tabular}
    }

 \vspace{-1.0em}
    \label{tab:ablation_graph}
\end{table*}

\begin{figure*}
\centering
\subfigure[Install]{
    \centering
    \includegraphics[width=0.48\linewidth]{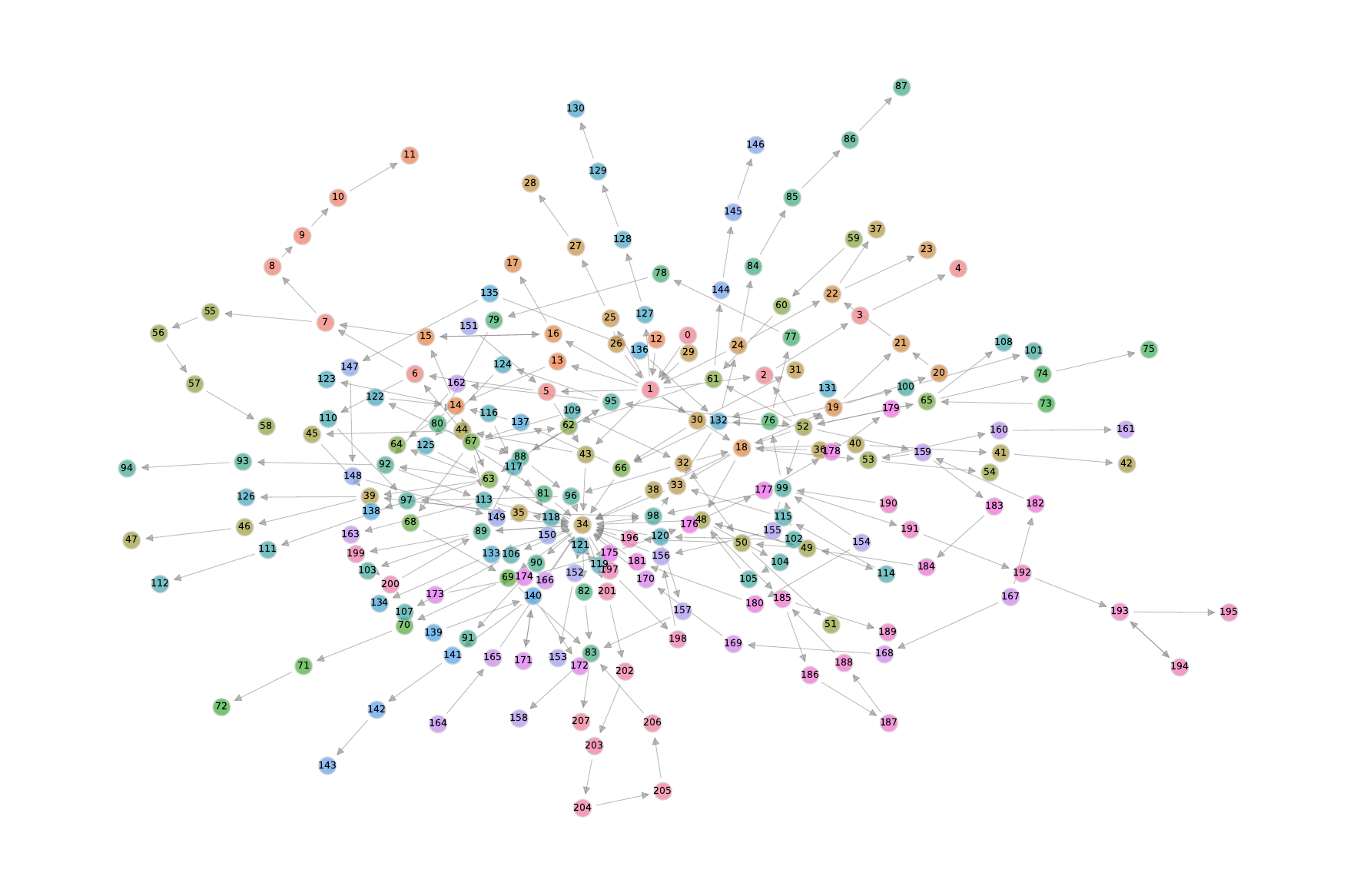}
  }
  \subfigure[WebShopping]{
    \centering
    \includegraphics[width=0.48\linewidth]{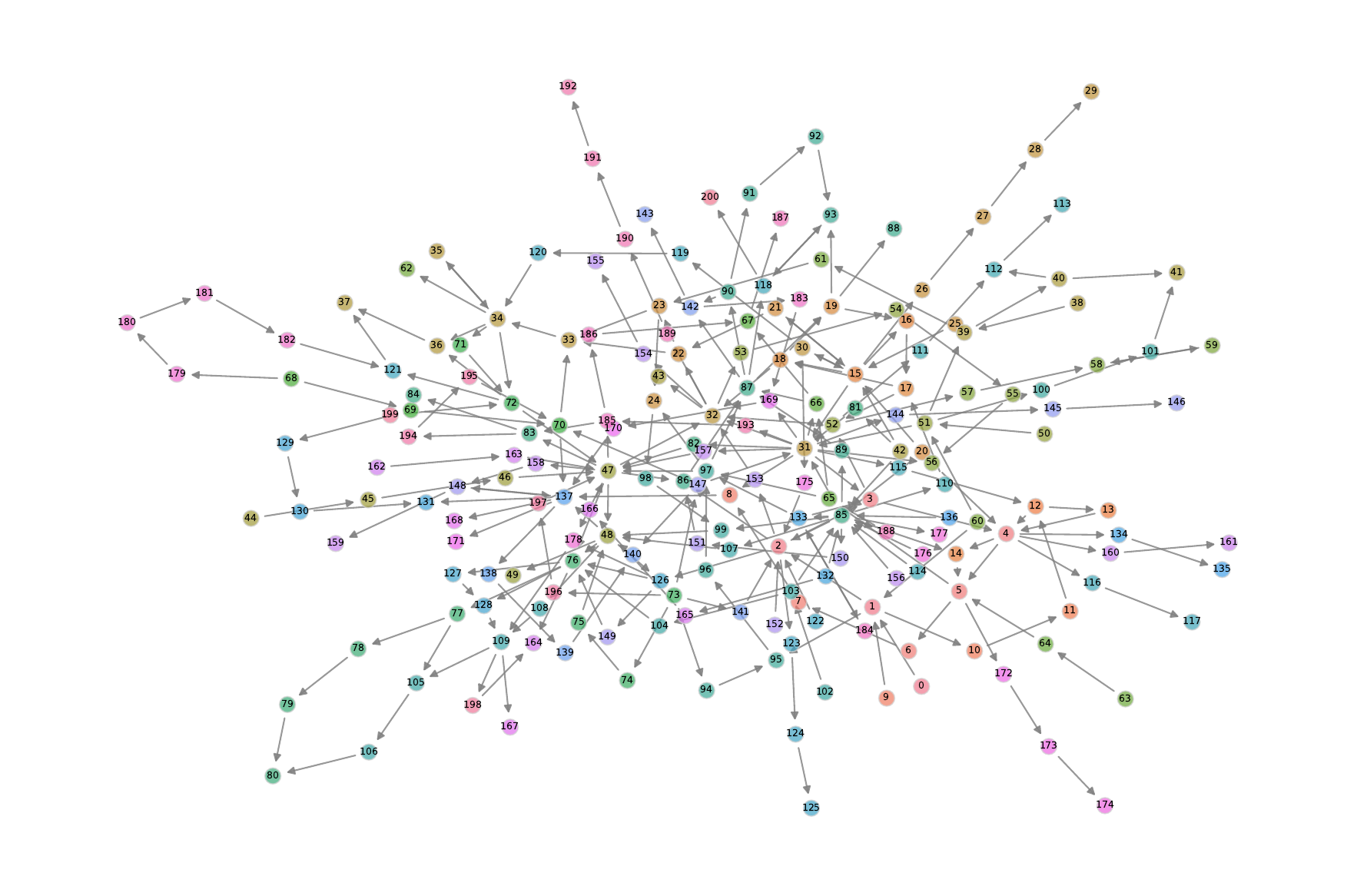}
  }
   \vspace{-0.8em}
  \caption{Examples of page graph visualizations of scenarios in AITW dataset.}

  \label{fig:graph_visual}
\end{figure*}

Meanwhile, introducing \textit{GL} to different agents also brings performance improvements. Furthermore, we find that the benefits of introducing \textit{GL} in the Decision Agent (D-Agent) are greater than those in the Sub-Task Planning Agent (STP-Agent). The same results are also observable in Table~\ref{tab:ablation_res_mind2web}, but in the web navigation tasks~\cite{mind2web}, agents have different preferences for \textit{GL} under different tasks; for example, in the Cross-Task and Cross-Domain split, introducing \textit{GL} to STP-Agent is better than D-Agent, but this result is reversed in the Cross-Website split. We attribute these results to differences in interaction logic across scenarios and varying navigation preferences of the base model for different devices.

To further validate \textit{GL}'s advantages, we conduct a fine-grained analysis of its impact on each decision step in the Mind2Web\cite{mind2web} dataset. As shown in Table~\ref{tab:ablation_action}, the introduction of \textit{GL} greatly improves the Opt.F1 score of the {\tt 'SELECT'} and {\tt 'TYPE'} actions. Regarding the decrease in performance of the {\tt 'CLICK'} action, we analyze the data and find that the reason is due to the inconsistency of the labels in the dataset itself; that is, the {\tt 'SELECT'} action has two label definitions at the same time: 1) two consecutive {\tt 'CLICK'} actions; 2) a single {\tt 'SELECT'} action. Our \model tends to choose more reasonable {\tt 'SELECT'} actions. However, this can lead to situations where it is judged as having failed, even when it executes the correct action. As a result, the Opt.F1 score for the {\tt 'SELECT'} type decreases.

\subsection{Page Graph Analysis}


\begin{table}[htbp]
\caption{Statistics of sampled episodes in different scenarios of dataset AITW and corresponding page graph.}
\centering
\resizebox{0.9\linewidth}{!}{
    \begin{tabular}{c|cccc}
    \toprule
    Scenario    & \# Episodes & \# Images & \# Nodes & \# Edges \\
    \midrule
    General     & 43          & 341        & 132      & 168      \\
    Install     & 55          & 538        & 208      & 286      \\
    G.Apps  & 24          & 198        & 67       & 93       \\
    Single      & 55          & 194        & 92       & 70       \\
    WebShop. & 56          & 712        & 201      & 323      \\
    \midrule
    Total       & 231         & 1983       & 700      & 940   
    \\
    \bottomrule
    \end{tabular}
    }
    \label{tab:aitw_graph_data}
\end{table}

For publicly available datasets (e.g., AITW, Mind2Web and GUI Odyssey), there are abundant data of episodes for the construction of page graphs, but it actually takes substantial costs for data collection. Therefore, in the previous evaluation, we only sample a small subset of episodes to build the graph, demonstrating the practicality of our framework. In this section, we use the full episodes from the dataset to build the graph and compare the result with Section~\ref{sec:main_res}. As shown in Table~\ref{tab:ablation_graph}, we find that \model utilizing full episodes only yields better performance in specific scenarios, where the random sampling version remains a competitive accuracy score. The results indicate that even if there are only limited episodes for reference, the page graph built from them can still provide effective guidance for \model.

To analyze the page graph deeply, we also collect the statistics of sampled episodes and the page graph. From Table~\ref{tab:aitw_graph_data}, we can find that the nodes in page graph is much less than the images of corresponding episodes, suggesting that there are lots of repeated pages in the same scenario. Meanwhile, the number of edges is also smaller than the number of actions(usually the same as number of images), which suggests than some consecutive in-page actions have been combined at one edge for the simplicity of the graph structure. More statistics on Mind2Web and GUI Odyssey datasets are listed in the Supplementary Materials.

\begin{figure}[tbp]
  \centering
   \includegraphics[width=1.0\linewidth]{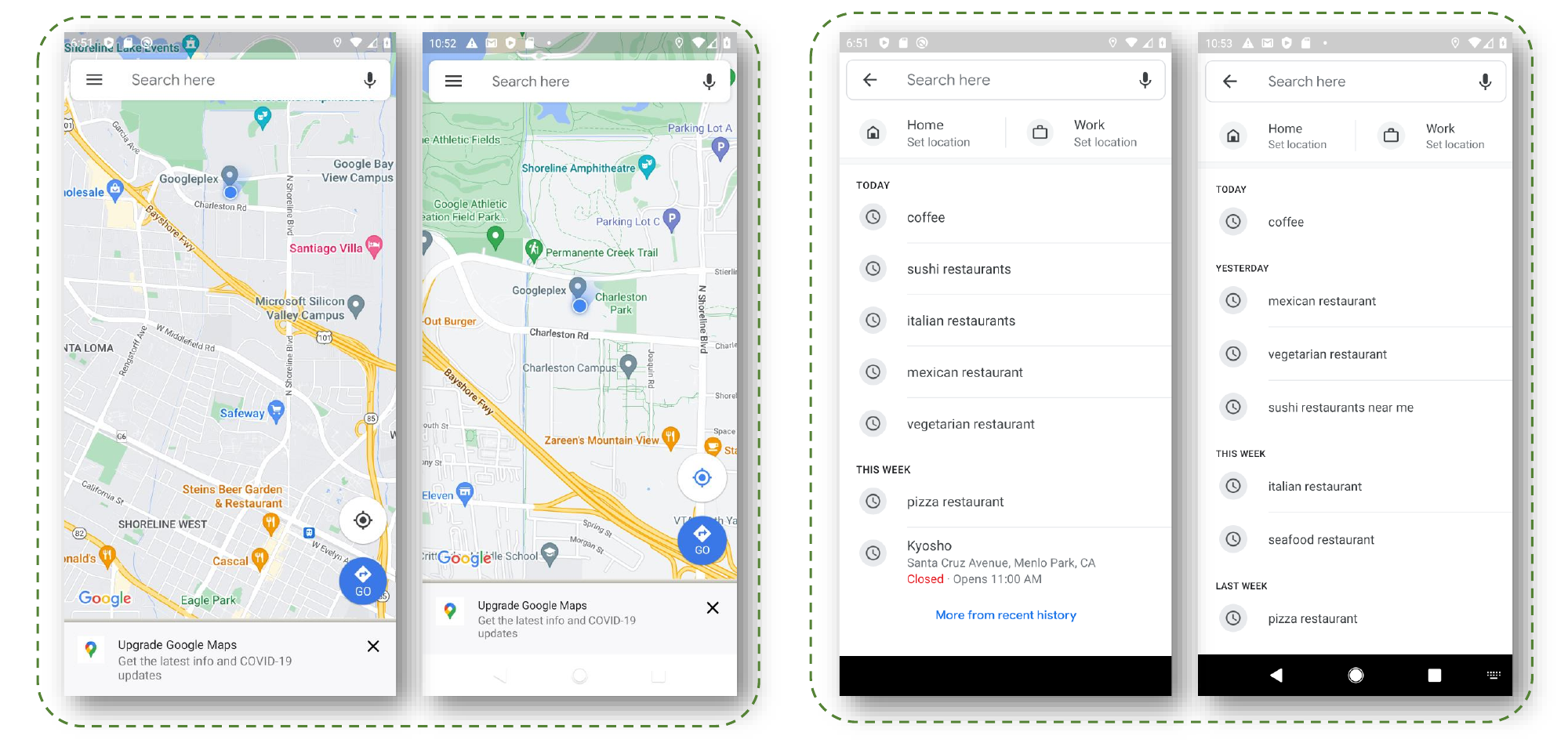}
\vspace{-1.5em}
   \caption{Cases of original images sharing the same node.}
    \vspace{-1.0em}
   \label{fig:repeat_case}
\end{figure}

Besides, we visualize some page graphs constructed by the episodes from different scenarios in AITW dataset. As shown in Figure~\ref{fig:graph_visual}, we can see that the in-degree or out-degree of some nodes in the graphs is greater than 1, indicating that some similar pages in the episodes share the same nodes. We also visualize some cases of these similar pages in Figure~\ref{fig:repeat_case}, demonstrating the effectiveness the dual-level similarity check.

\section{Conclusion}

In this paper, we design an automated pipeline to reconstruct the discrete chained episodes into the page graph, capturing the complex transition relationships between screen pages. To fully utilize this prior knowledge as the perceptions of GUI scenarios, we further propose a tailored multi-agent framework \model equipped with the RAG technology to retrieve perception guidelines from the graph to improve the planning and execution process. Extensive experiments on three benchmark datasets illustrate the effectiveness of \model powered by page graphs, even when the available episodes are limited.

\begin{acks}
This work was supported by the National Natural Science Foundation of China (Grant No. 62372408, 62476245). This work was supported by Ant Group Research Fund.
\end{acks}

\bibliographystyle{ACM-Reference-Format}
\bibliography{sample-base}

\appendix






\section{Structural details of page graph.}
In this section, we show the details of the page graphs of Mind2Web (Table~\ref{tab:mind2web_graph_data}) and GUI Odyssey (Table~\ref{tab:odyssey_graph_data}) generated using the page graph construction pipeline. "$\#$ Episodes" represents the number of episodes we sampled from the original data. "$\#$ Images" is the number of screenshots in the sampled data. "$\#$ Nodes" and "$\#$ Edges" denote the number of nodes and edges in the page graph after pipeline processing, i.e., page jump determination, node similarity check and page graph update (Algorithm~\ref{alg:framework}).

\begin{table}[h]
\centering
\caption{Statistics of sampled episodes in different scenarios of dataset Mind2Web and corresponding page graph.}
\resizebox{0.9\linewidth}{!}{
    \begin{tabular}{c|cccc}
    \toprule
    Scenario    & \# Episodes & \# Images & \# Nodes & \# Edges \\
    \midrule
    Entertainment & 63   & 342  & 113   & 95    \\
    Travel        & 122  & 1001 & 172   & 159   \\
    Shopping      & 67   & 484  & 131   & 111   \\
    \midrule
    Total      & 252  & 1827 & 416   & 365    
    \\
    \bottomrule
    \end{tabular}
    }
    
    \label{tab:mind2web_graph_data}
\end{table}

\begin{table}[h]
\centering
\caption{Statistics of sampled episodes in different scenarios of dataset GUI-Odyssey and corresponding page graph.}
\resizebox{\linewidth}{!}{
    \begin{tabular}{c|cccc}
    \toprule
    Scenario    & \# Episodes & \# Images & \# Nodes & \# Edges \\
    \midrule
    Tool        & 25   & 311  & 119   & 157   \\
    Information       & 17   & 314  & 96    & 133   \\
    Shopping        & 8    & 144  & 58    & 66    \\
    Media       & 16   & 166  & 60    & 85    \\
    Social      & 15   & 210  & 72    & 100   \\
    Multi-Apps  & 35   & 736  & 210   & 388   \\
    \midrule
    Total      & 116  & 1881 & 615   & 929    
    \\
    \bottomrule
    \end{tabular}
    }
    
    \label{tab:odyssey_graph_data}
\end{table}

\section{Guidance statistics on each benchmark.}
\begin{figure}[H]
  \centering
   \includegraphics[width=0.9\linewidth] 
   {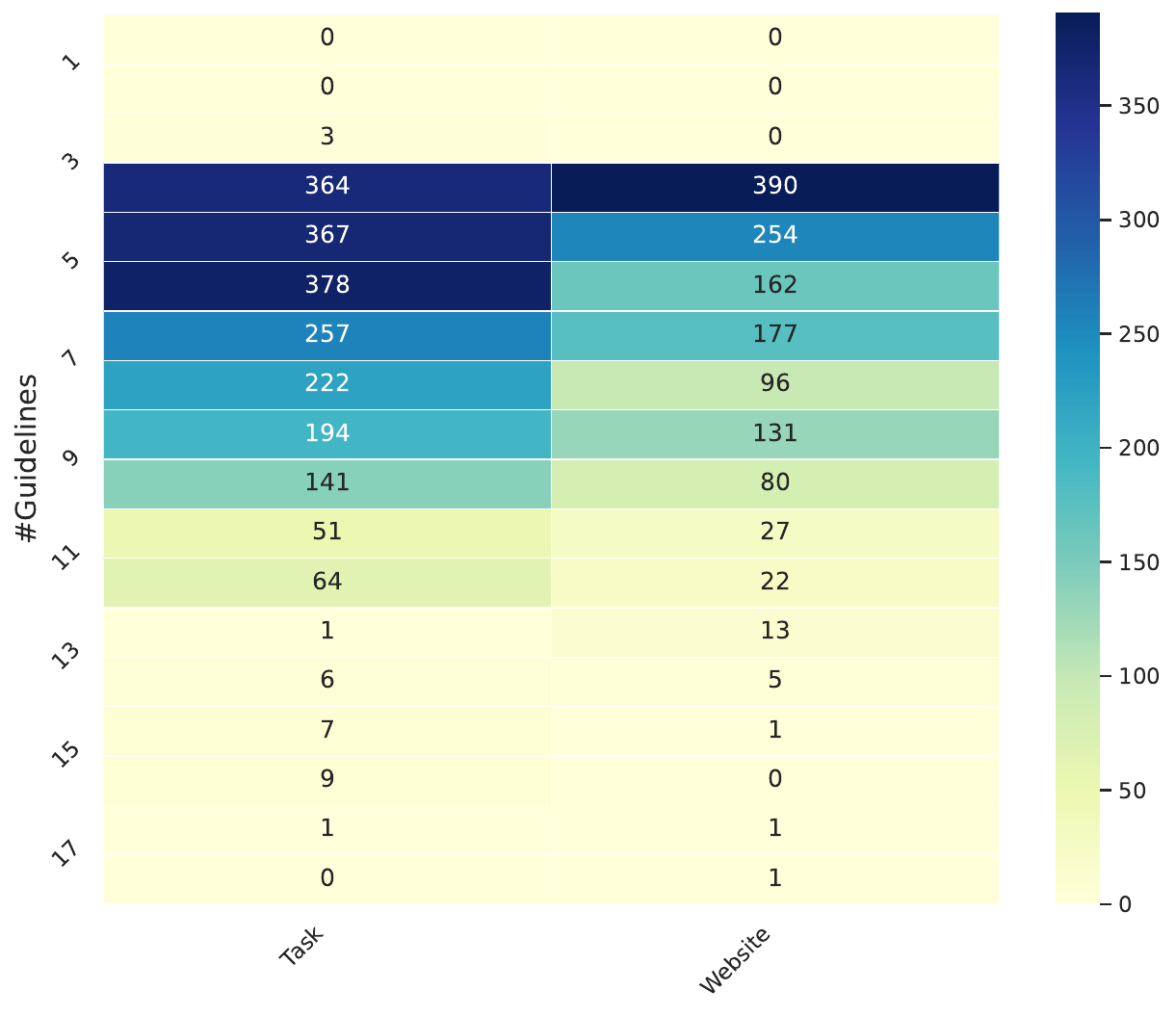}
   \caption{The data distribution of guidelines in Mind2Web dataset.}
   \label{fig:ref_mind2web}
\end{figure}

\begin{figure}[H]
  \centering
   \includegraphics[width=.9\linewidth] 
   {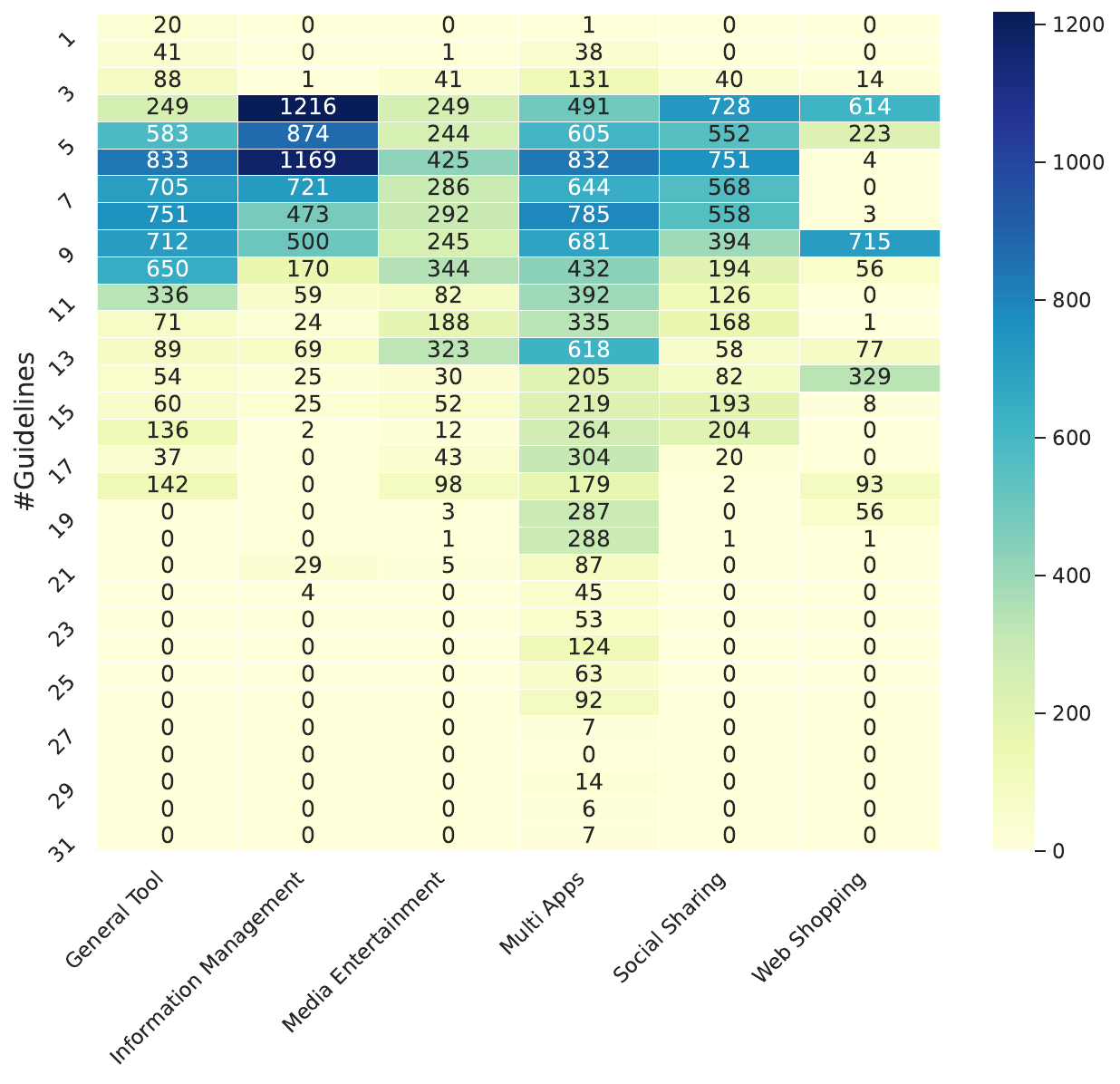}
   \caption{The data distribution of guidelines in GUI Odyssey dataset.}
   \label{fig:ref_odyssey}
\end{figure}

In our \model, guidelines (\textit{GL}) represent GUI knowledge retrieved from the prior knowledge base, i.e., page graph (Section~\ref{page_graph_construction}), and are used to enhance the agent’s decision making in GUI navigation flow. Here we exhibit the more \textit{GL} distribution in Figure~\ref{fig:ref_mind2web} (Mind2Web) and Figure~\ref{fig:ref_odyssey} (GUI Odyssey) show the other two benchmarks details of retrieved GLs. It can be seen that for different datasets, our RAG strategy can effectively retrieve enough GLs to assist the agent's action decision.


\begin{figure*}[]
    \centering
    \includegraphics[width=1.0\linewidth] 
    {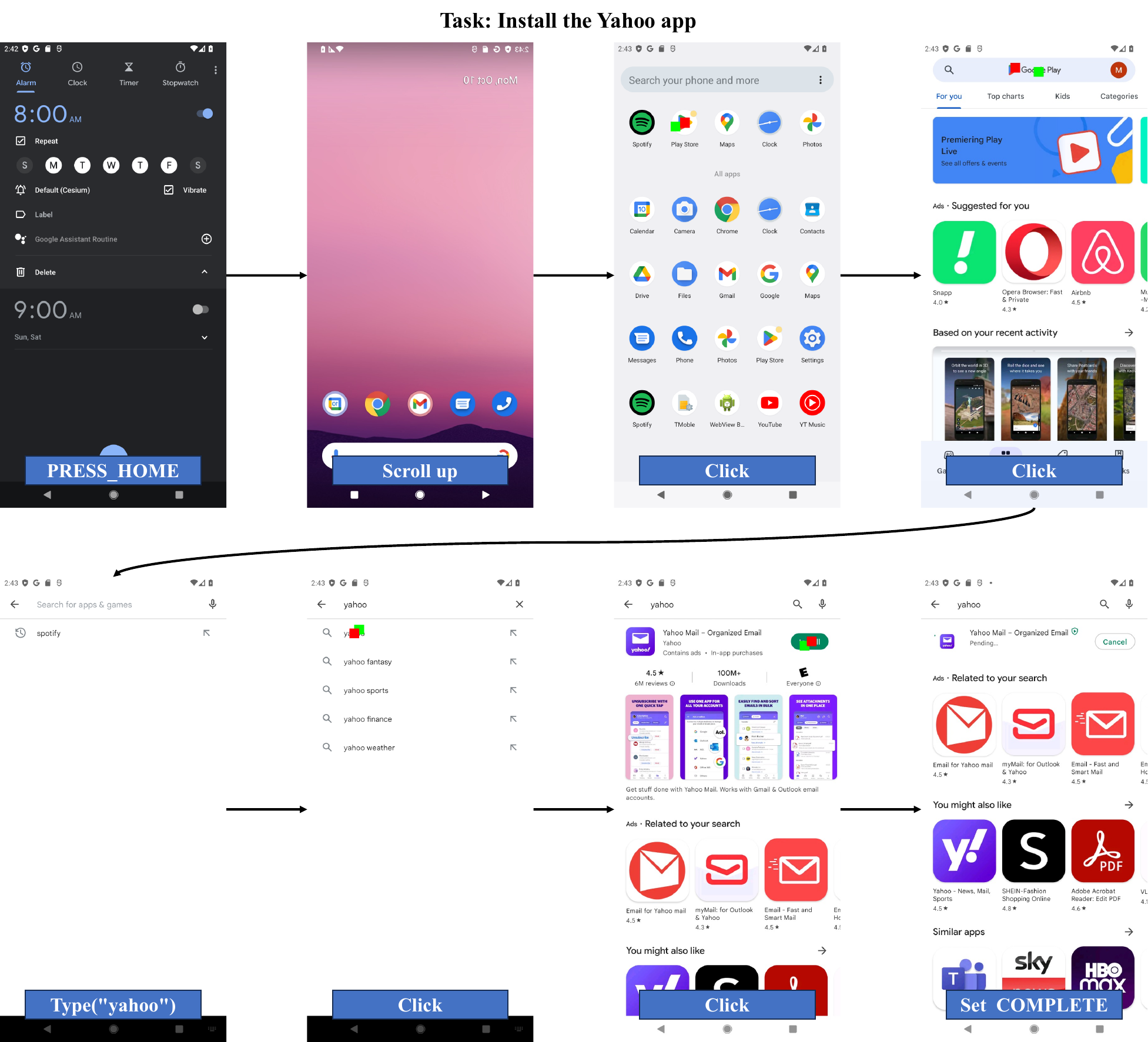}
    \caption{Navigation process of \model.}
    \label{fig:good_case}
\end{figure*}

\begin{figure*}[]
  \centering
  \subfigure[]{
    \centering
    \includegraphics[width=0.7\linewidth]{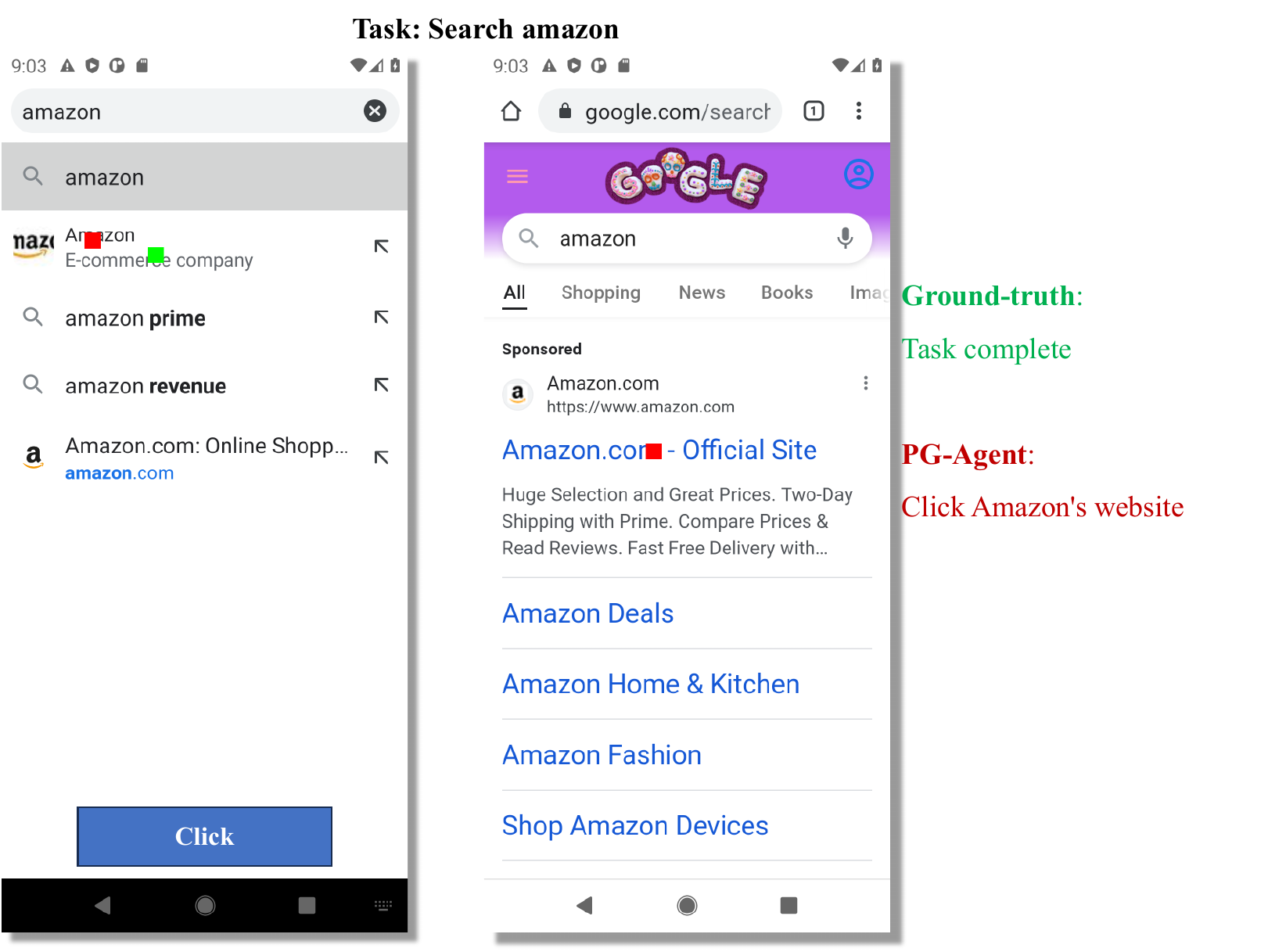}
  }
  \subfigure[]{
    \centering
    \includegraphics[width=\linewidth]{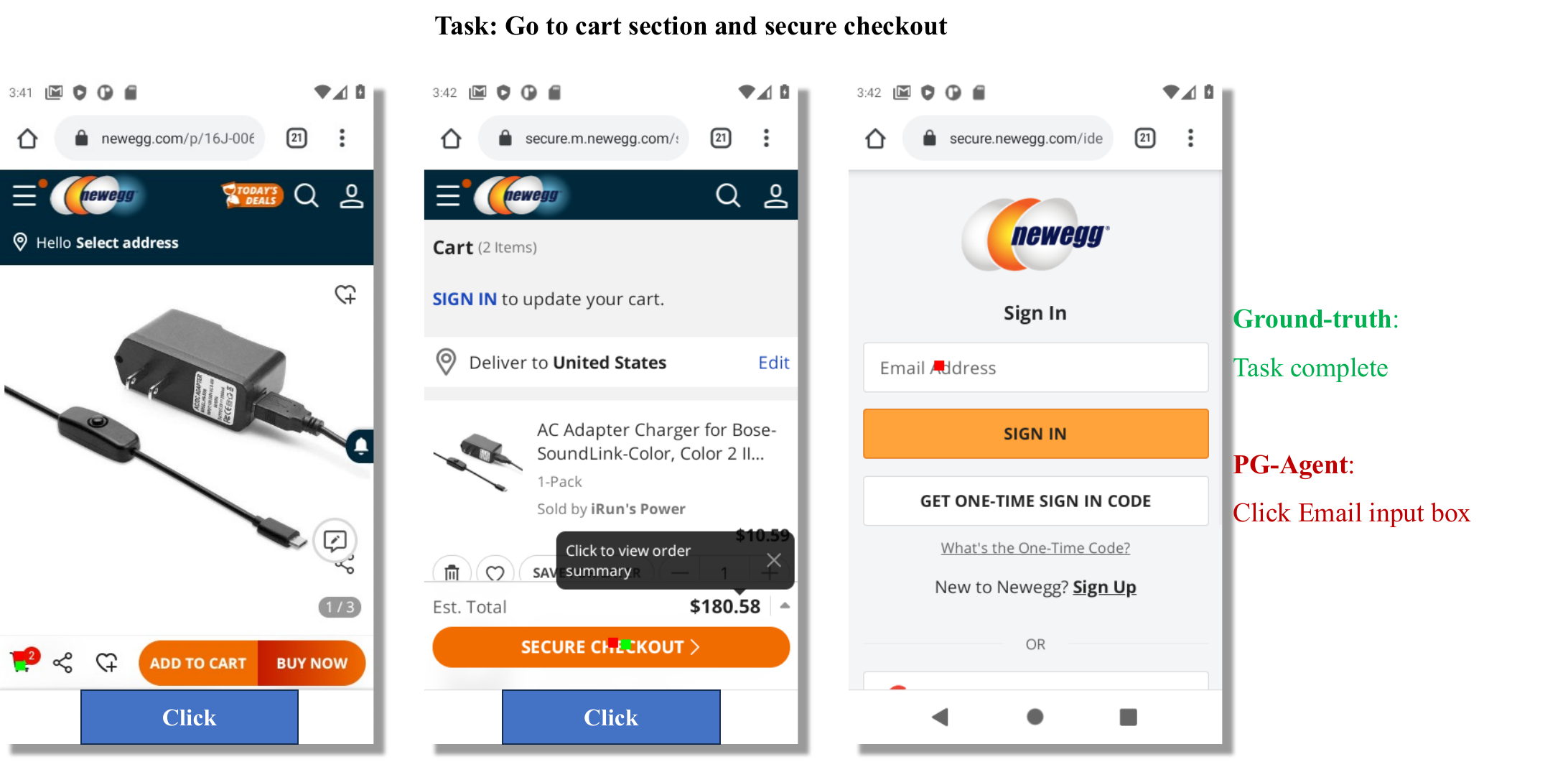}
  }
  \caption{Cases determined as failure steps in "Single" scenario.}
  \label{fig:bad_case}
\end{figure*}
 
\section{Case Study}
In this section, we select a part of cases for detailed analysis. When facing the Click operation, we use red rectangle to mark ground-truth, and green rectangle to mark the location where \model clicks.

As shown in Figure~\ref{fig:good_case}, our \model can successfully complete tasks with long steps.
At the same time, we figure out that the inconsistency between ground-truth and the model's judgment on when the task should be completed led to some failed cases. In Figure~\ref{fig:bad_case}(a), the task is to search Amazon. After getting the search results of Amazon, \model believes that it is necessary to click Amazon's website to complete the whole task. A similar situation occurs in Figure~\ref{fig:bad_case}(b). When secure checkout is performed, the pop-up sign in interface naturally means that the task is not over yet, further sign in is required to complete. This is exactly what \model thinks, which is inconsistent while the answer determines the task is already completed.

\section{Pseudocode of page graph construction and multi-agent workflow.}
In this part, we present the details of two core modules in the form of pseudocode, \textit{(i)} the generation pipeline of the page graph (Algorithm~\ref{alg:framework}) and \textbf{(ii)} the workflow of the Agent (Algorithm \ref{alg_multi_agent_flow}). Through Algorithm~\ref{alg:framework}, we can automatically transform discrete chain-like episodes into high-quality page graph as GUI prior knowledge base. Then, with the empowerment of page graph, the multi-agent system as Algorithm~\ref{alg:framework} can effectively complete the GUI navigation task.

\begin{algorithm}[ht]

    \renewcommand{\algorithmicrequire}{\textbf{Input:}}  
    \renewcommand{\algorithmicensure}{\textbf{Output:}} 
    \caption{The pipeline of page graph construction.}  
    \label{alg:framework}
    \begin{algorithmic}[1]
        \REQUIRE Actions $A$;Images $I$;Image Locations $L$;Task $T$.
        \ENSURE Page graph $\mathcal{G}$.
        \STATE $N_{before}\gets \varnothing$
        \STATE $Q_{action}\gets [ ]$
        \STATE $\mathcal{G}\gets [ ]$
        \FOR{$i \in 1,2,...|E|$}        
        \STATE \textcolor{gray}{// Page Jump Determination} 
        \IF{$i > 1$}
        \STATE $S_{action}^{(i-1)}\gets \{I^{(i-1)},A^{(i-1)}\}$
        \STATE $Q_{action} \gets Q_{action} \cup \{S_{action}^{(i-1)}\}$
        \STATE $Y_{jump}^{(i)}\gets \{I^{(i-1)},I^{(i)},S_{action}^{(i-1)}\}$
        \IF{$Y_{jump}^{(i)}$=\text{'No'}}
        \STATE Continue
        \ENDIF
        \ENDIF
        
        \STATE \textcolor{gray}{// Node Similarity Check}
        \STATE $S_{page}^{(i)} \gets I^{(i)}$
        \STATE $S_{node}^{(i)} \gets \{S_{page}^{(i)},\mathcal{G}\}$
        \STATE $id \gets \{I^{(i)},S_{node}^{(i)}\}$
        \STATE $I_{id},N_{id} \gets \{id,L,\mathcal{G}\}$
        \STATE $Y_{dissimilar}^{(i)}\gets \{I^{(i)},I_{id}\}$
        
        \STATE \textcolor{gray}{// Page Graph Update}
        \IF{$Y_{dissimilar}^{(i)}$=\text{'Yes'}}
        \STATE $N_{new}\gets\{I^{(i)},L^{(i)}\}$
        \ELSE 
        \STATE $N_{new}\gets N_{id}$
        \ENDIF
        \IF{$i > 1$}
        \STATE $E_{new}\gets\{Q_{action},T\}$
        \STATE $\mathcal{G}=\mathcal{G} \cup (N_{before},E_{new},N_{new})$
        \ENDIF
        \STATE $N_{before}\gets N_{new}$
        \ENDFOR
        
        \RETURN $\mathcal{G}$
    \end{algorithmic}  
\end{algorithm}

\begin{algorithm}[ht]
    \renewcommand{\algorithmicrequire}{\textbf{Input:}}   
    \renewcommand{\algorithmicensure}{\textbf{Output:}} 
    \caption{The workflow of \model.}  
    \label{alg_multi_agent_flow}
    \begin{algorithmic}[1]
        \REQUIRE user's goal $T_g$; current screen state $I_t$; Maximum length of episode $H$; page graph $\mathcal{G}$; Observation Agent $\mathcal{O}_{agent}$; Global Planning Agent $\mathcal{P}^G_{agent}$; Sub-Task Planning Agent $\mathcal{P}^S_{agent}$; Decision Agent $\mathcal{D}_{agent}$.
        \ENSURE Action decision $\mathcal{R}_d$ based on current screen state $I_t$.
        
        \STATE\tt \textcolor{gray}{// Guidelines Retrieval}
        \STATE $G_{I_t}\gets RAG(I_t, \mathcal{G})$
        \STATE\tt \textcolor{gray}{// Task Decomposition}
        \STATE $\mathcal{R}_g \gets \mathcal{P}^G_{agent}(I_t,T_g)$
        \STATE $t\gets0$
        \STATE $\tau\gets[]$ 
        \WHILE{$t<H$ and $\mathcal{R}_d \neq$ "COMPLETE"}
        
        \STATE\tt \textcolor{gray}{// Observation Generation}
        \STATE $\mathcal{R}_o \gets \mathcal{O}_{agent} (I_t,T_g,\tau_{<t})$

        \STATE\tt \textcolor{gray}{// Candidate Action Generation}
        \STATE $\mathcal{R}_s\gets \mathcal{P}^S_{agent}(I_t,\mathcal{R}_o,\mathcal{R}_g,G_{I_t},\tau_{<t})$
        \STATE\tt \textcolor{gray}{// Final Decision}
        \STATE $\mathcal{R}_d\gets \mathcal{D}_{agent} (I_t,\mathcal{R}_o,\mathcal{R}_s,G_{I_t},\tau_{<t})$
        \STATE $\tau \gets \tau \cup \mathcal{R}_d$
        \STATE $t\gets t+1$ 
        \ENDWHILE
    \end{algorithmic}
\end{algorithm}




\end{document}